\documentclass[preprint,12pt,authoryear]{elsarticle}

\usepackage{amssymb}

\usepackage[OT1]{fontenc}
\usepackage{amsthm,amsmath}
\usepackage{setspace}
\usepackage[authoryear]{natbib}
\usepackage{graphicx}
\usepackage{url} 
\usepackage{enumerate}
\usepackage{rotating}

\journal{Physica A: Statistical Mechanics and its Applications}

\begin{document}

\begin{frontmatter}



\title{Robust multivariate and functional archetypal analysis with application to financial time series analysis \tnoteref{label1}}
\tnotetext[label1]{{The code and data for reproducing the examples  are available at \url{www3.uji.es/~epifanio/RESEARCH/rofada.rar}. A preliminary version of this work was presented at the 8th  International Conference on Mathematical and Statistical Methods for Actuarial Sciences and Finance (MAF 2018) (\cite{MolEpi18}), where the application data were analyzed in a non-robust way.} }

\author[label2]{Jes\'us Moliner}
\address[label2]{Dept. Matem\`atiques. Campus del Riu Sec. Universitat Jaume I, 12071 Castell\'o, Spain}

\author[label2,label3]{Irene Epifanio\corref{cor1}}
\address[label3]{Institut de Matem\`atiques i Aplicacions de Castell\'o.}
\cortext[cor1]{Tel.: +34 964728390; fax: +34 964728429.}

\ead{epifanio@uji.es}

\begin{abstract}
Archetypal analysis approximates data by means of mixtures of actual extreme cases (archetypoids) or archetypes, which are a convex combination of cases in the data set. Archetypes lie on the boundary of the convex hull. This makes the analysis very sensitive to outliers. A robust methodology by means of M-estimators for classical multivariate and functional data is proposed. This unsupervised methodology allows  complex data to be understood even by non-experts. The performance of the new procedure is assessed in a simulation study, where a comparison with a previous  methodology for the multivariate case is also carried out, and our proposal obtains  favorable results. Finally, robust bivariate functional archetypoid analysis is applied to a set of companies  in the S\&P 500 described by two time series of stock quotes. A new graphic representation is also proposed to visualize the results.  The analysis shows how the information can be easily interpreted and how even non-experts can gain a qualitative understanding of the data.
\end{abstract}

\begin{keyword}
Multivariate functional data \sep Archetype analysis \sep Stock \sep M-estimators \sep Multivariate time series


\end{keyword}

\end{frontmatter}


\section{Introduction}
\label{sec:intro}  Many econometric data are big data in the form of time series that can be seen as functions (\cite{Tsay16}). 
In Functional Data Analysis (FDA) the observations are functional time series or multivariate functions. \cite{Ramsay05}  provide an excellent overview of FDA. FDA has been applied in many different fields (\cite{Ramsay02}), and although it is a relatively new area of research in the business and economic sectors,  applications are beginning to proliferate in these fields (\cite{ASMB:ASMB388,doi:10.1080/07350015.2015.1092976,doi:10.1080/07350015.2017.1279058}). 

Data mining of functional time series (\cite{FU2011164}) is as important as in the classical multivariate version. It is desirable to  understand and describe the entire data set and to be able to extract information that is easily interpretable even by non-experts. We deal with an unsupervised statistical learning problem since only input features and not output are present. Data decomposition techniques  to  find the latent components are usually employed in classical multivariate statistics (see \cite[Chapter 14]{HTF09} for a complete review of unsupervised learning techniques). A data matrix is considered as a linear combination of several factors. The constraints on the factors and how they are combined  give rise to different unsupervised techniques (\cite{Morup2012,Thurau12,Vinue15}) with different goals. For instance, Principal Component Analysis (PCA) explains data variability  adequately at the expense of the interpretability of the factors, which is not always straightforward since  factors are linear combinations of the features. For their part,  clustering techniques such as $k$-means or $k$-medoids return factors that are easily interpreted. Note that data are explained through several centroids, which are means of groups of data in the case of $k$-means, or medoids, which are concrete observations in the case of $k$-medoids. Nevertheless,  the binary assignment of data to the clusters diminishes their modeling flexibility as compared with PCA. 

Archetype analysis (AA) lies somewhere in between these two methods, as the interpretability of its factors is as easy as with clustering techniques but  its modeling flexibility is higher than for clustering methodologies. \cite{Vinue15} provide a summary table showing the relationship between several multivariate unsupervised techniques, as do \cite{Morup2012}. \cite{STONE1996110} also showed that AA may be more appropriate than PCA when the data do not have elliptical distributions.

AA was formulated by \cite{Cutler1994}. In AA each observation in the data set is approximated by a mixture (convex combination) of pure or extremal types called archetypes. Archetypes themselves are restricted to be convex combinations of the individuals in the data set. However,  AA may not be satisfactory in some fields since, being artificial constructions, nothing guarantees the existence of subjects in our sample with  characteristics similar to those of the archetypes (\cite{Seiler2013}).
In order to solve this issue, the new concept of archetypoids was introduced by  \cite{Vinue15}. In Archetypoid Analysis (ADA)  each observation in the data set is approximated by a convex combination of a set of real extreme observations called archetypoids. AA and ADA were extended to dense functional data by \cite{Epifanio2016} and to sparse functional data by \cite{VinEpi17}. In the functional context, functions from the data set, which can be multivariate functions,  are explained by mixtures of archetypal functions.

This process not only allows us to relate the subjects of the sample to extreme patterns but also facilitates comprehension of the data set. Humans understand the data better when the individuals are shown through their extreme constituents (\cite{Davis2010}) or when features of an individual are shown as opposed to those of another (\cite{Thurau12}). In other words, as regards human interpretability, the central points returned by clustering methods do not seem as good as extreme types, which are also more easily understandable than a linear combination of data.

 AA and ADA have therefore aroused the interest of researchers working in different fields, such as astrophysics (\cite{Chan2003}), biology (\cite{Esposito2012}), climate (\cite{doi:10.1175/JCLI-D-15-0340.1,w9110873,EpiIbSi18}), developmental
psychology (\cite{SAM16}), e-learning (\cite{Theodosiou}), engineering (\cite{EpiVinAle,EpiIbSi17,MillanEpi}), genetics (\cite{Morup2013}), machine learning (\cite{Morup2012,EugsterPAMI,Eugster14,Rago15}), multi-document summarization (\cite{Canhasi13,Canhasi14}), nanotechnology (\cite{doi:10.1021/acsnano.5b05788}), neuroscience (\cite{griegos,Morup16}) and sports (\cite{Eugster2012}). AA has also been applied in market research (\cite{Li2003,Porzio2008,Midgley2013}) in the multivariate context. Despite the fact that financial time series are commonly analyzed by unsupervised techniques ranging from PCA (\cite{Alexander, Tsay,Ingrassia2005}) to clustering (\cite{DOSE2005145,BASALTO2007635,TSENG201211,DURSO20132114,DIAS2015852,AnnMaharaj2010,CAPPELLI201323,doi:10.1002/cem.2565,DURSO201233,doi:10.1142/S0218488504002849,1516151,Alonso2018}), including robust versions of these (\cite{6737103,Verdonck,DURSO20161,DURSO201812}),  to the best of our knowledge functional archetypal analysis has not been used in financial or business applications until now.


As archetypes are situated on the boundary of the convex hull of data (\cite{Cutler1994}), AA and ADA solutions are sensitive to outliers.
The problem of robust AA in the multivariate case was addressed by \cite{Eugster2010}. The idea is to find the archetypes of the large majority of the data set rather than of the totality. \cite{Eugster2010} considered a kind of M-estimators for multivariate real-valued  data ($m$-variate), where the domain of their loss function is not $\mathbb{R}^+$, but $\mathbb{R}^m$. Recently, \cite{Sinova}  considered functional M-estimators for the first time, where the domain of their loss function is $\mathbb{R}^+$. We base our proposal to robustify archetypal solutions in the multivariate and functional case on this last kind of loss function, which is commonly used in robust analysis (\cite{maronna}). 
    
The main novelties of this work consist of: 1. Proposing a robust methodology for classical multivariate and functional AA and ADA; 2. Introducing a new visualization procedure that makes it easy to summarize  the results and multivariate time series; 3. Applying  functional archetypal analysis  to financial time series for the first time, more specifically  to multivariate financial time series. 

Section \ref{metodologia} reviews AA and ADA for the multivariate and functional case and Section \ref{metodologiarobusta} introduces their respective robust versions. Our proposal is compared with a previously existing methodology for robust multivariate AA in Section \ref{simulacion}, where a simulation study with functional data is also carried out to validate our procedure. In Section \ref{resultados}, robust ADA is applied to a data set of 496 companies that are characterized
by two financial time series. Furthermore, some visualization tools are also introduced in this section. Finally,  conclusions
and future work are discussed in Section \ref{conclusiones}. The code in R (\cite{R}) and data for reproducing the results are available at \url{www3.uji.es/~epifanio/RESEARCH/rofada.rar}.

\section{Archetypal analysis}
\label{metodologia}
\subsection{AA and ADA in the classical multivariate case} 
Let $\mathbf{X}$ be an $n \times m$ matrix with $n$ cases and $m$ variables. In AA, three matrices are sought: a) the $k$ archetypes $\mathbf{z}_j$, which are the rows of a $k \times m$ matrix $\mathbf{Z}$; b) an $n \times k$ matrix $\mathbf{\alpha} = (\alpha_{ij})$ that contains the mixture coefficients that approximate each case $\mathbf{x}_i$ by a mixture of the archetypes  ($\mathbf{\hat{x}}_i = \displaystyle \sum_{j=1}^k \alpha_{ij} \mathbf{z}_j$); and c) a $k \times n$ matrix $\mathbf{\beta} = (\beta_{jl})$ that contains the mixture coefficients that define each archetype ($\mathbf{z}_j$ = $\sum_{l=1}^n \beta_{jl} \mathbf{x}_l$). To determine these matrices, the following residual sum of squares (RSS) with the respective constraints is minimized ($\| \cdot\|$ denotes the Euclidean norm for vectors):

\begin{equation} \label{RSSar}
RSS = \displaystyle  \sum_{i=1}^n \| \mathbf{x}_i - \sum_{j=1}^k \alpha_{ij} \mathbf{z}_j\|^2 = \sum_{i=1}^n \| \mathbf{x}_i - \sum_{j=1}^k \alpha_{ij} \sum_{l=1}^n \beta_{jl} \mathbf{x}_l\|^2{,}
\end{equation}

under the constraints

\begin{enumerate}

\item[1)] $\displaystyle \sum_{j=1}^k \alpha_{ij} = 1$ with $\alpha_{ij} \geq 0$ {for} $i=1,\ldots,n$ {and}

\item[2)] $\displaystyle \sum_{l=1}^n \beta_{jl} = 1$ with $\beta_{jl} \geq 0$ {for} $j=1,\ldots,k${.}

\end{enumerate}

It is important to mention that archetypes do not necessarily match real cases. Specifically, this will only happen when one and only one $\beta_{jl}$ is equal to one for each archetype, i.e. when each archetype is composed of only one case that presents the entire weight. Therefore, in ADA the previous constraint 2) is changed by the following one, and as a consequence the previous continuous optimization problem of AA is transformed into a mixed-integer optimization problem:

\begin{enumerate}

\item[2)] $\displaystyle \sum_{l=1}^n \beta_{jl} = 1$ with $\beta_{jl} \in \{0,1\}$ and $j=1,\ldots,k$.
\end{enumerate}

\cite{Cutler1994} demonstrated that archetypes are located on the boundary of the convex hull of the data if $k$ $>$ 1, although this does not necessarily happen for archetypoids (see \cite{Vinue15}). However, if $k$ = 1, the archetype coincides with the mean and the archetypoid with the medoid (\cite{Kaufman90}).

\cite{Cutler1994} developed  an alternating minimizing algorithm to estimate the matrices in the AA problem. It alternates between calculating the best $\mathbf{\alpha}$ for given archetypes $\mathbf{Z}$ and the best archetypes $\mathbf{Z}$ for a given $\mathbf{\alpha}$. In each step a 
penalized version of
the non-negative least squares algorithm by \cite{Lawson74} is used to solve the convex least squares problems. That algorithm, with certain modifications (previous data standardization and use of  spectral norm in equation \ref{RSSar} instead of  Frobenius norm for matrices), was implemented by \cite{Eugster2009} in the R package {\bf archetypes}. In our R implementation those modifications were canceled and the data are not standardized by default and the objective function to optimize coincides with equation \ref{RSSar}.

As regards the estimation of the matrices in the ADA problem, \cite{Vinue15} developed an algorithm based on the idea of the Partitioning Around Medoids (PAM) clustering algorithm (\cite{Kaufman90}). This algorithm consists of two stages: the BUILD phase and the SWAP phase. In the BUILD phase, an initial set of archetypoids is computed, while that set is improved during the SWAP phase by exchanging the chosen observations for unselected cases and by inspecting whether these replacements diminish the RSS. \cite{JSSv077i06} implemented that algorithm in the R package {\bf Anthropometry} with three possible initial sets in the BUILD step.  One of them is referred to as the $cand_{ns}$ set and  consists of
 the nearest neighbors in Euclidean distance to the $k$ archetypes. The second candidates, the {$cand_{\alpha}$} set,  consist of the observations with the maximum $\alpha$ value for each archetype $j$, i.e. the observations with the largest relative share for the respective archetype. The third initial candidates, the so-called {$cand_{\beta}$} set, are the cases with the maximum $\beta$ value for each archetype $j$, i.e. the cases that most influence  the construction of the archetypes. Each of these three sets goes through the SWAP phase and three sets are obtained. From these three sets, the one with lowest RSS (often the same set is retrieved from the three initializations) is selected as the ADA solution.

 Archetypes are not necessarily nested and neither are archetypoids. Therefore, changes in $k$ will yield different conclusions. This is why the selection criterion is particularly important. Thus,  if the researcher has prior knowledge of the structure of the data, the value of $k$ can be selected based on that information. Otherwise, the elbow criterion, which has been widely used (\cite{Cutler1994,Eugster2009,Vinue15}), could be considered. With the elbow criterion, the RSS is represented for different $k$ values and the value $k$ is chosen as the point where the elbow is found.

\subsection{AA and ADA in the functional case}
In FDA each datum is a function. In this context, the values of the $m$ variables in the classical multivariate context become function values with a continuous index $t$, and the data set adopts the form $\{x_1(t),...,x_n(t)\}$ with $t \in [a,b]$.
It is assumed that these functions belong to a Hilbert space, they satisfy reasonable smoothness conditions and are square-integrable functions on that interval. In addition, in the definition of the inner product, the sums are replaced by integrals. 

Again, the goal of functional archetype analysis (FAA) is to approximate the functional data sample by mixtures of $k$ archetype functions. 
 The difference with the multivariate case is that now both archetypes and observations are functions. In FAA, two matrices $\alpha$ and $\beta$ are also calculated to minimize the RSS. However, certain aspects should be highlighted. On the one hand, RSS are now calculated with a functional norm (the $L^2$-norm, $\|f\|^2= <f,f> = \int_a^b f(t)^2 dt$, is considered) instead of a vector norm. On the other hand, observational and archetype vectors $\mathbf{x}_i$ and $\mathbf{z}_i$ now correspond to observational and archetype functions $x_i(t)$ and $z_i(t)$. In any case, the interpretation of matrices $\alpha$ and $\beta$ is the same as in the standard multivariate case.

Functional archetypoid analysis (FADA) is also an adaptation of ADA, changing vectors for functions. In this regard, FADA  aims to find $k$ functions of the sample (archetypoids) that approximate the functions of the sample through the mixtures of these functional archetypoids. Again, vector norms are replaced by functional norms.  Interpretation of the matrices is the same as before.

In practice, the functions are recorded at discrete times. Standard AA and ADA could be applied to the function values of $m$ equally-spaced values from $a$ to $b$ to obtain FAA and FADA. However, this approach is not computationally efficient (\cite{Epifanio2016}). Therefore, we represent data by means of basis functions. This reduces noise, i.e.  functions are smoothed. Furthermore,  data observations do not have to be equally spaced,  the number of observed points can vary across records and they can be measured at different time points. This also makes it possible  to perform a more efficient analysis, since the number of coefficients of the basis functions is usually smaller than the number of  time points evaluated. 
 The crux of the matter is to choose an appropriate basis together with the number of basis elements. Nevertheless, this issue appears repeatedly  in all FDA problems. Functions of the sample should be expanded by basis functions that share common features (see \cite{Ramsay05} for a detailed explanation about smoothing functional data). For densely observed functions, the case that concerns us, the basis coefficients are computed separately for each function, while information from all functions should be used to calculate the coefficients for each function (\cite{James}) for sparsely observed functions. 

Let us see how the RSS is formulated in terms the coefficients $\mathbf{b}_i$, the vector of length $m$ that approximates $x_i(t)$ $\approx$ $\sum_{h=1}^m b_i^h B_h(t)$ with the basis functions  $B_h$  ($h$ = 1, ..., $m$) (see \cite{Epifanio2016} for details):

\begin{equation}\label{RSSfar}
\begin{split}
 RSS = \displaystyle \sum_{i=1}^n \| {x}_i - \sum_{j=1}^k \alpha_{ij} {z}_j\|^2 = \sum_{i=1}^n \| {x}_i - \sum_{j=1}^k \alpha_{ij} \sum_{l=1}^n \beta_{jl} {x}_l\|^2 = 
 \sum_{i=1}^n {\mathbf{a}}'_i \mathbf{W} {\mathbf{a}}_i
 {,}
\end{split}
\end{equation}
where $\mathbf{a'}_i$ = $\mathbf{b'}_i - \sum_{j=1}^k \alpha_{ij} \sum_{l=1}^n \beta_{jl} \mathbf{b'}_l$ and $\mathbf{W}$ is the order $m$ symmetric matrix with the inner products of the pairs of basis functions $w_{m_1,m_2}$ = $\int B_{m_1}B_{m_2}$. If the basis is orthonormal,  for instance the Fourier basis, $\mathbf{W}$ is the order $m$ identity matrix and FAA and FADA can be computed  using AA and ADA with the basis coefficients. If not, $\mathbf{W}$ has to be computed previously one single time by numerical integration.

\begin{figure}
\begin{center}
\begin{tabular}{ccc}
	\includegraphics[width=.3\linewidth]{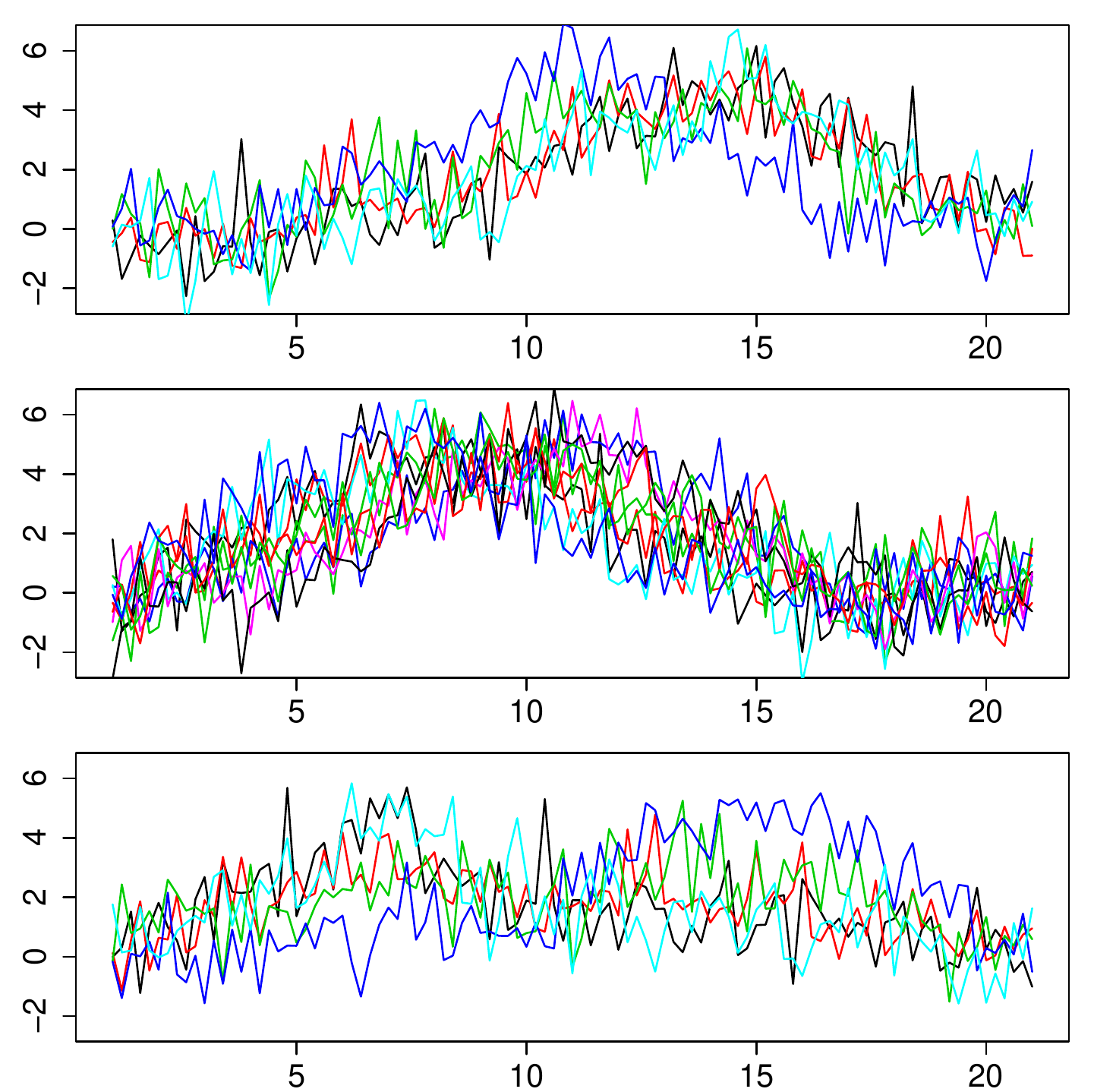} & \includegraphics[width=.3\linewidth]{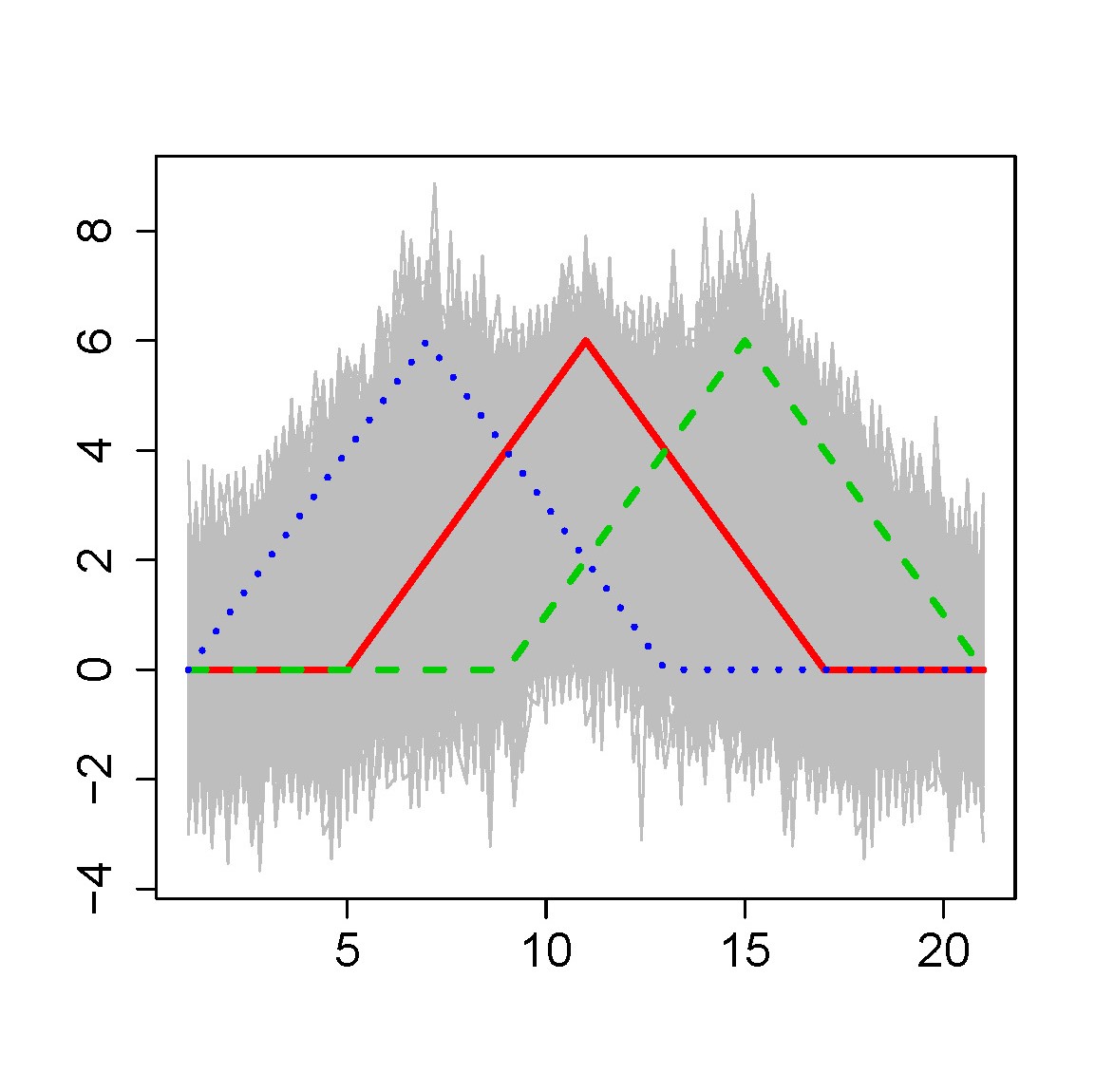} & 
	\includegraphics[width=.3\linewidth]{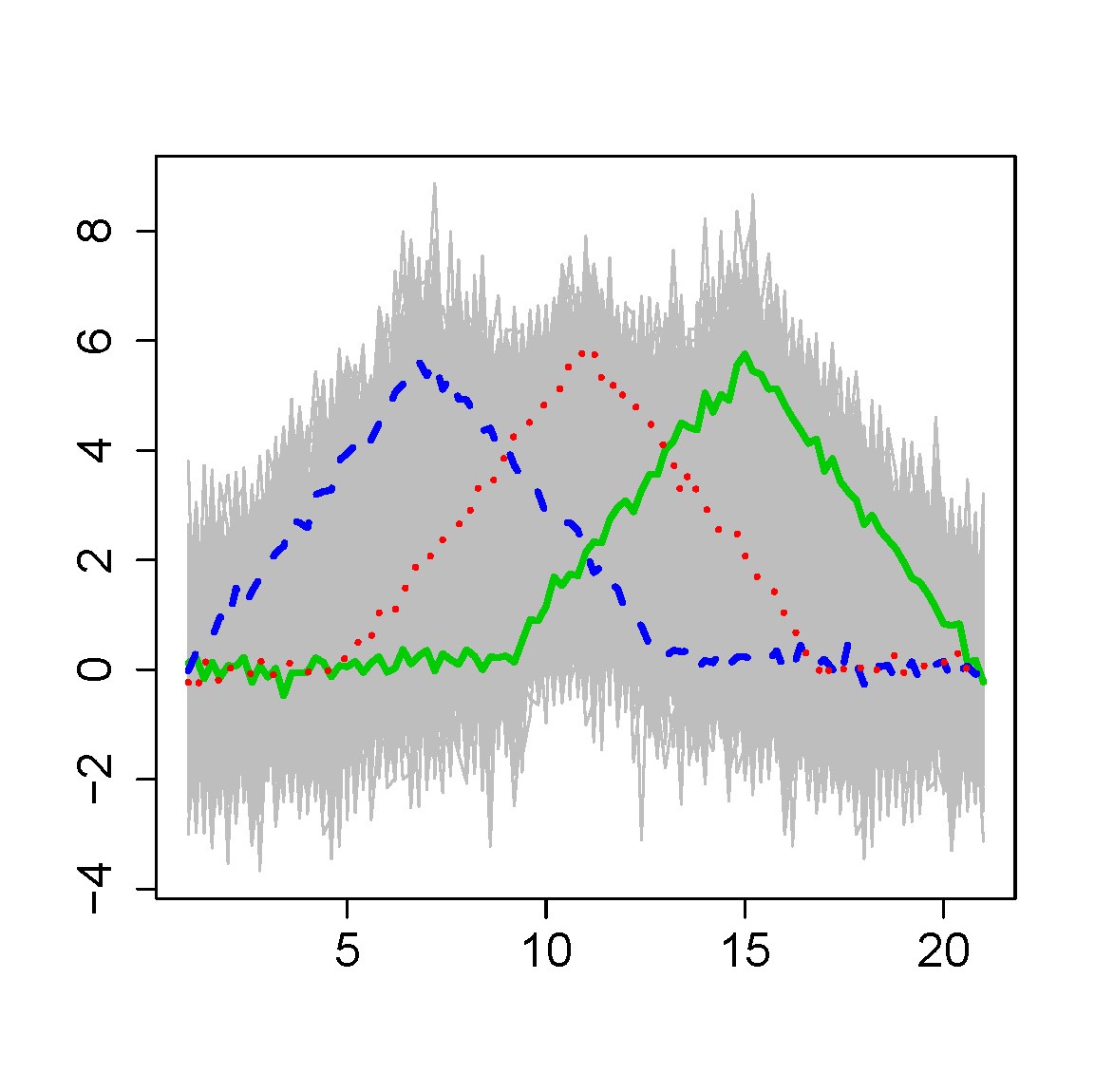} \\
	(a) & (b) & (c)\\
	\includegraphics[width=.3\linewidth]{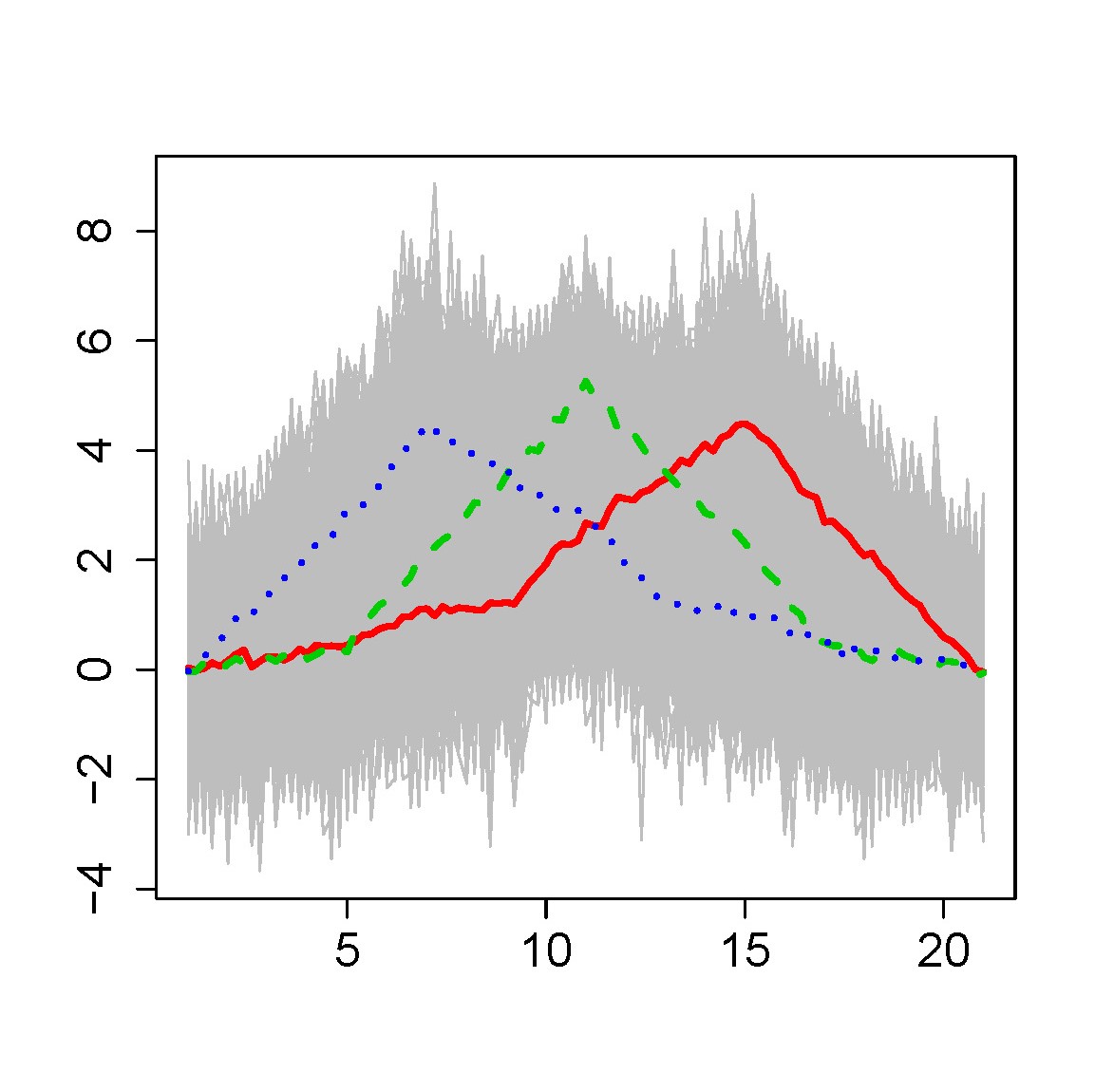} & \includegraphics[width=.3\linewidth]{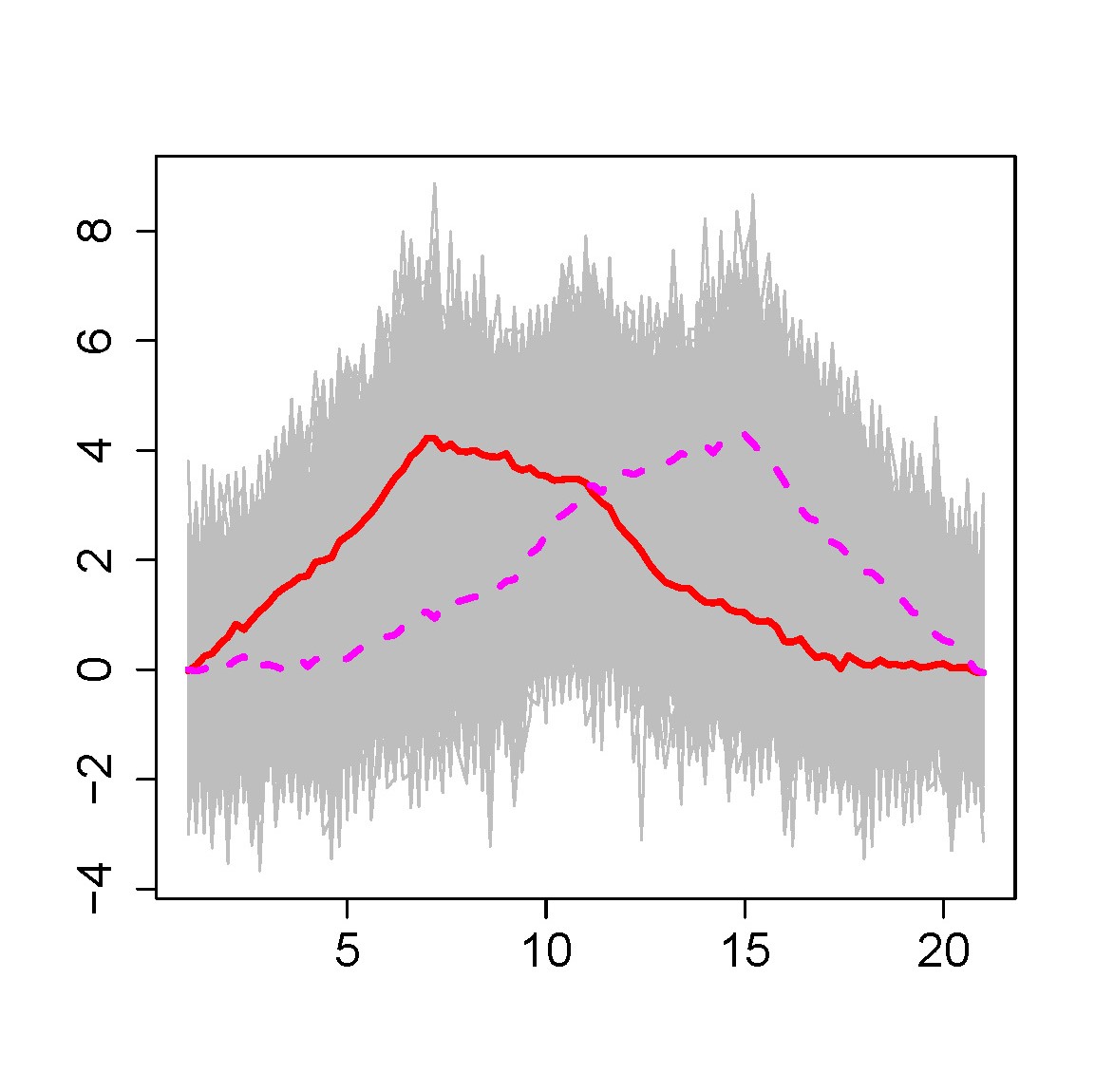} & 
	\includegraphics[width=.3\linewidth]{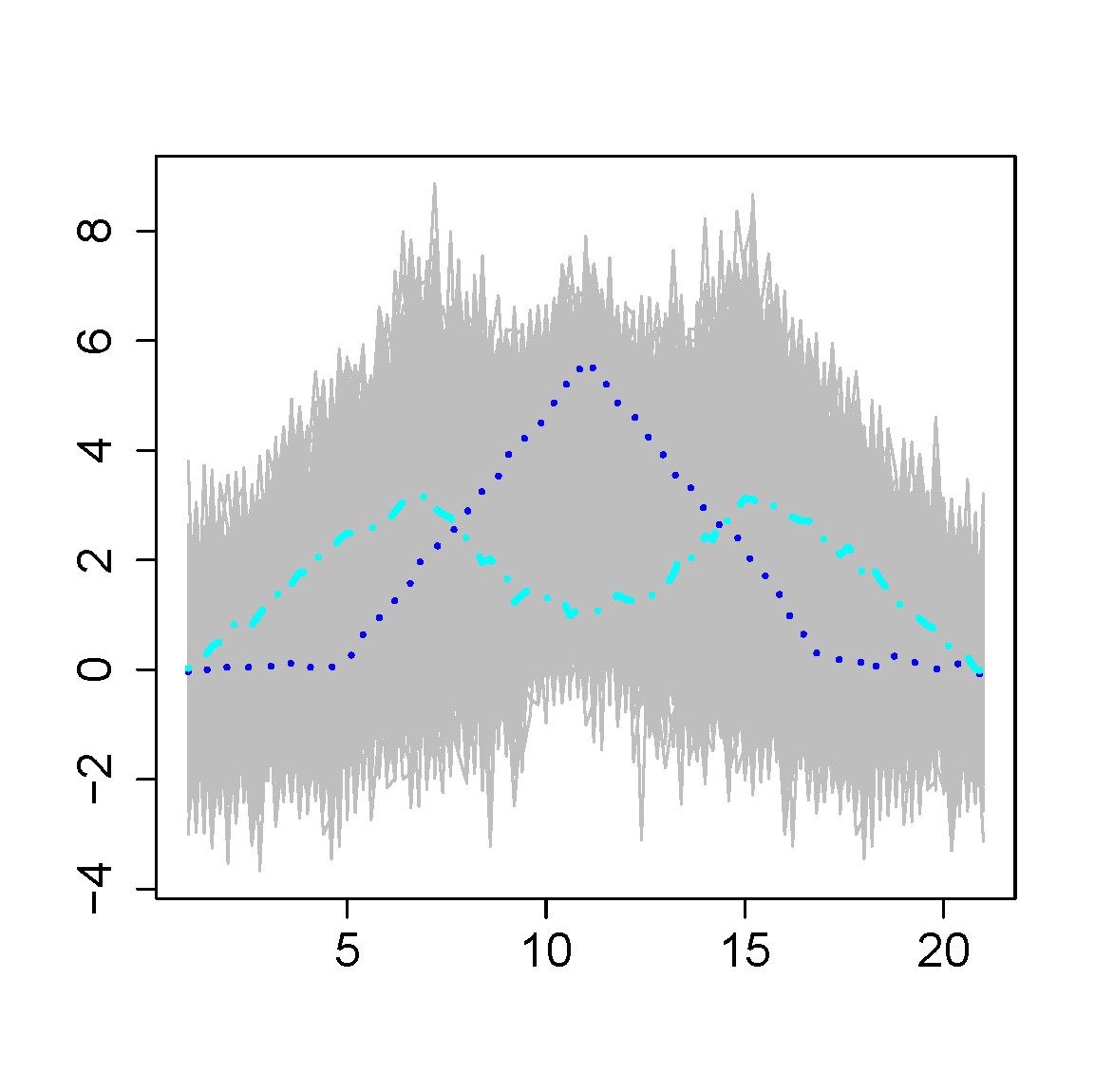} \\
	(d) & (e) & (f)
	\end{tabular}
	\end{center}
	\caption{{Waveform. (a) Five functions for each of the three
classes, respectively. (b) $h_j$ functions together with the data set in gray. (c) Archetypes. (d) Centers of $k$-means. (e) Effect of adding and subtracting a multiple of PC 1 to the mean curve. (f) Effect of adding and subtracting a multiple of PC 2 to the mean curve. }
	\label{waveform}}
\end{figure}

{Let us see a toy example to illustrate what archetypes mean and the differences compared with PCA and clustering. We use a functional version, previously considered by \citet{Ferraty} and \citet{Epifanio08}, of the well-known simulated
data known as waveform data \citep{Breiman}. Functions $x$ are discretized
 at 101
points ($t$ = 1, 1.2, 1.4, ..., 21) such that
\begin{itemize}
    \item $x(t) = uh_1(t) + (1 - u)h_2(t) + \epsilon(t)$ for class 1,
    \item
 $x(t) = uh_1(t) + (1
- u)h_3(t) + \epsilon(t)$ for class 2, and
    \item $x(t) = uh_2(t) + (1 - u)h_3(t) + \epsilon(t)$ for class 3,
\end{itemize}
where $u$ is uniform on $[0, 1]$, $\epsilon(t)$ are standard
normal variables, and  $h_i$ are the shifted triangular
waveforms: $h_1(t)$ = $max(6 - |t - 11|, 0)$,  $h_2(t)$ = $h_1(t -
4)$ and $h_3(t)$ = $h_1(t + 4)$. Note that $x$ are mixtures of $h_j$ ($j$ = 1, 2, 3), and therefore $h_j$ are archetype functions by definition. The toy example has 150 waveforms in each of the 3 classes. Figure \ref{waveform} shows the simulated data set, together with the results for AA, PCA and clustering. Note that archetypes are estimations of $h_j$ functions, the purest profiles, unlike cluster centers, which are not so extreme and whose profiles are not as clear as those of archetypes, since they are central points: the mean of each class. PC 1 quantifies a discriminability trade-off between classes 1 and 2, whereas PC 2 quantifies a discriminability trade-off between class 3 and the rest of the classes. In summary, finding
extreme profiles, which are easily interpretable, is not the objective of clustering or PCA, but
it is the intention of AA and ADA, together with the expression of the data as a mixture of those extreme profiles.}

\subsubsection{Multivariate functional archetypal analysis} 
It is common to analyze data with more  than one dimension. In our context, this means working with samples in which we analyze more than one function for each individual, so each function describes a characteristic of the subject. 

First of all, we need to define an inner product between multivariate functions. The simplest definition is to add up the inner products of the multivariate functions. Therefore, the squared norm of a $P$-variate function is the sum of the squared norms of the $P$ components. Consequently, FAA and FADA for $P$-variate functions is equivalent to $P$ independent FAA and FADA with shared parameters $\alpha$ and $\beta$.
In practical terms, this means to work  with a composite function formed by stringing the $P$ functions together.

Without loss of generality, let $f_i(t)=(x_i(t),y_i(t))$ be a bivariate function. So, its squared norm is $\|f_i\|^2= \int_a^b x_i(t)^2 dt + \int_a^b y_i(t)^2 dt$. Let $\mathbf{b}_i^x$ and $\mathbf{b}_i^y$ be the vectors of length $m$ of the coefficients for $x_i$ and $y_i$  for the basis functions $B_h$. Therefore, to compute FAA and FADA, the RSS is reformulated as:

\begin{equation}\label{RSSfarb}
\begin{gathered}
RSS ={} \sum_{i=1}^{n}||f_i- \sum_{j=1}^{k} \alpha_{ij}z_j||^2 
= \sum_{i=1}^{n}||f_i- \sum_{j=1}^{k} \alpha_{ij}\sum_{l=1}^{n}\beta_{jl}f_l||^2 
= \sum_{i=1}^{n}||x_i- \sum_{j=1}^{k} \alpha_{ij}\sum_{l=1}^{n}\beta_{jl}x_l||^2 \\ + \sum_{i=1}^{n}||y_i- \sum_{j=1}^{k} \alpha_{ij}\sum_{l=1}^{n}\beta_{jl}y_l||^2 
=  \sum_{i=1}^{n}{\mathbf{a}^x}'_i \mathbf{Wa}_i^x+\sum_{i=1}^{n}{\mathbf{a}^y}'_i \mathbf{Wa}_i^y
\end{gathered}
\end{equation}
where $\mathbf{{a^x}}'_i={\mathbf{b}^x}'_i - \sum_{j=1}^{k} \alpha_{ij}\sum_{l=1}^{n}\beta_{jl} {\mathbf{b}^x}'_l $ and  ${\mathbf{a}^y}'_i={\mathbf{b}^y}'_i - \sum_{j=1}^{k} \alpha_{ij}\sum_{l=1}^{n}\beta_{jl} \mathbf{{b^y}}'_l $ with the corresponding AA or ADA constraints for $\mathbf{\alpha}$ and $\mathbf{\beta}$. The union of $\mathbf{b}^x_i$ and $\mathbf{b}^y_i$ composes the observations. If the basis functions are orthonormal, FAA and FADA  are reduced to computing standard AA and ADA for the $n \times 2m$ coefficient matrix composed by joining the coefficient matrix for $x$ and $y$ components.

\section{Robust archetypal analysis} \label{metodologiarobusta}
The RSS is formulated as the sum of the squared  (vectorial or functional) norm of the residuals, $r_i$ ($i$ = 1, ..., $n$). Here, $r_i$ denote vectors of length $m$ in the multivariate case or (univariate or multivariate) functions in the functional case. The least squares loss function does not provide robust solutions since it favors outliers; large residuals have large effects. M-estimators try to lower the large influence of outliers by changing the square loss
function for a less rapidly increasing loss function.  \cite{Eugster2010} defined a loss function from $\mathbb{R}^m$ to $\mathbb{R}$. However, \cite{Sinova} established several conditions of the loss function $\rho$ for functional M-estimators, the first of which is that the loss function is defined from $\mathbb{R}^+$ to $\mathbb{R}$ and the loss is specified as $\rho(||r_i||)$. {Furthermore, $\rho(0)$ should be zero, $\rho(x)/x$ should tend towards zero, when $x$ tends towards zero, and $\rho$ should be differentiable and $\rho'$ and $\phi(x)$ = $\rho'(x)/x$ should be both continuous and
bounded, where we assume that $\phi(0)$ := $lim_{x \rightarrow 0}$ $\rho'(x)/x$ exists and is finite. This last condition is not satisfied by the standard least squares loss function $\rho(x)$ = $x^2$ ($\rho'$ is not bounded). Details about properties of  functional M-estimators, such as their consistency and robustness  by means of
their breakdown point and their influence function can be found in \cite{Sinova}.}

\cite{Sinova} also analyzed the common families of loss functions. We follow the ideas of \cite{Sinova} and the Tukey biweight or bisquare family of loss function (\cite{10.2307/1267936}) with tuning parameter $c$ is considered, since this loss function copes  with  extreme outliers well. Therefore, RSS in equations \ref{RSSar}, \ref{RSSfar} and \ref{RSSfarb} are replaced by $\sum_{i=1}^n \rho_c(||r_i||)$, where $\| \cdot\|$ denotes the Euclidean norm for vectors, the $L^2$-norm for univariate functions and corresponding norm for $P$-variate functions, and $\rho_c(||r_i||)$ is given by

\begin{equation} \label{robustrho}
  \rho_c(||r_i||) =\left\{ \begin{array}{ll}
            c^2/6 \cdot (1 - (1 - ||r_i||^2/c^2)^3) & \mbox{if 0 $\leq$ $||r_i||$ $\leq$ $c$}\\
            c^2/6 & \mbox{if $c$ $<$ $||r_i||$}
            \end{array}
						\right.
            \end{equation}

For the tuning parameter $c$, we follow \cite{Cleveland}, as did \cite{Eugster2010}, and $c$ = 6$me$ with $me$ being the median of the residual norms unequal to zero{, although other alternatives are analyzed in the simulation study}. From the computational point of view, we only have to replace RSS with this new objective function in the previous algorithms. It only depends on the norm of the residuals, so in the functional case, it can be expressed in terms of the coefficients in the basis and $\mathbf{W}$, and no integration is needed. 

{Note that $\rho_c$ is not scale equivariant, i.e. the results depend heavily on the units of measurement, which is why in both the real case \citep{maronna} and the functional case \citep{Sinova} the tuning parameter should take into account the
distribution of the data (the residuals in our case), in particular certain percentile of this distribution. Theoretically, the tuning parameter should be chosen such that the loss function is well adapted to the data \citep{Sinova}, but in practice there is no one gold standard method for selecting the tuning parameter.} 

Another possibility would be to use the Huber family of loss functions (\cite{huber:1964}) that depends on a tuning parameter. For example, this loss function was used by \cite{chen:hal-00995911} {and \cite{SUN2017147}, where} the tuning parameter was manually set and no suggestion was given about its selection. An important difference between Huber and the bisquare family is that residuals larger than $c$ contribute the same to the loss in this last family, which is not the case with the Huber family. For that reason, the bisquare family can better cope with extreme outliers.

\section{Simulation study}
\label{simulacion}
\subsection{Multivariate data} To compare the performance of our proposal, the same procedures and data set, known as ozone, examined by  \cite{Eugster2010} are considered. The data, which are also used as a demo in the R library {\bf archetypes}, consist of 330 observations of 9 standardized variables that are related to air pollution. We create a corrupted data set by adding a total of 5 outliers, as  \cite{Eugster2010} did. Table \ref{percentiles} shows the percentiles of 3 archetypes (A1, A2 and A3) extracted in four situations: a) original AA with the  original data set before adding the outliers (AAO), this solutions plays the gold standard reference role; b) original AA with the corrupted data set (AAC); c) robust AA by \cite{Eugster2010} with the corrupted data set (RAA-EL); d) our proposal of robust AA with the corrupted data set (RAA-ME). 

At a glance it can be seen that the archetypes returned by our proposal are the most similar to the original ones. To corroborate it, the Frobenius norm of the difference between the gold standard reference and the different alternatives applied to the corrupted data set has been computed with the following results for each situation: AAC  206.3; RAA-EL 221.3; RAA-ME  64.5. Our proposal provides the solution with the smallest difference with respect to the gold standard reference, while the robust AA solution by \cite{Eugster2010} provides a greater difference than the non-robust AA solution. For both AAC and  RAA-EL one of the archetypes (A2) is built entirely of a mixture of the added outliers, which is not the case with our proposal (0.07 is the only $\beta$ weight for the outliers). Therefore, a more robust solution is achieved with our proposal.

\begin{table}  
\caption{Percentiles profiles in four situations for the ozone data set (see text for details). \label{percentiles}}
\centering
\begin{tabular}{rrrrrrrrrrrrr}
Situation & \multicolumn{3}{c}{AAO} & \multicolumn{3}{c}{AAC} & \multicolumn{3}{c}{RAA-EL} & \multicolumn{3}{c}{RAA-ME}\\
Variable & A1 & A2 & A3 & A1 & A2 & A3 & A1 & A2 & A3 & A1 & A2 & A3\\ \hline 
OZONE&12&89&12&3&100&70& 12&99&81&3&97&12\\
500MH& 3&92&65&6&100&54&7&99&73&3&99&50\\
WDSP& 96&43&5&63&100&27& 43&99&27&78&78&8\\
HMDTY&56&82&11&19&100&45& 20&99&56&45&97&11\\
STMP&4&93&22&5&100&66& 10&99&80&5&98&16\\
INVHT&100&14&63&99&100&4& 70&99&5&99&41&56\\
PRGRT&92&64&2&36&100&36& 34&99&40&79&78&2\\
INVTMP&1&93&49&5&100&75&9&99&84&3&97&40\\
VZBLTY&77&15&77&87&100&9& 76&99&9&76&38&76
\end{tabular}
\end{table}

{We have also analyzed the influence that the tuning parameter has on the results. Instead of $c$ = 6$me$, we consider the following alternatives: $c$ = $P_j$, with $j$ = 25, 50 and 75, representing the 25th, 50th and 75th percentiles of the residual norms unequal to zero, and the same but multiplied by 6, i.e. $c$ = 6 $P_j$. Table \ref{cnorm} shows the Frobenius norm of the difference between the gold standard reference and RAA-ME computed using the different values of $c$. The same results are obtained if we use $c$ = 6$P_{25}$ and our selection,  $c$ = 6$P_{50}$, but they are worse if $c$ = 6$P_{75}$ is used; even so the result continues to be better than that for RAA-EL and similar to AAC. A slight improvement is achieved using $c$ = $P_{75}$, but if we use $c$ = $P_{25}$ or $c$ = $P_{50}$, the results are worse, although better than those obtained by using AAC and RAA-EL. So, except for one case, $c$ = 6$P_{75}$, where the results are similar to those without robustifying, we obtain more robust results for all the $c$ considered.}

\begin{table}  
\caption{{Frobenius norm of the difference between AAO and RAA-ME for different $c$.} \label{cnorm}}
\centering
\begin{tabular}{lrrr}
Multiplicative & \multicolumn{3}{c}{Percentile}\\
Factor & 25th & 50th & 75th\\ \hline 
1 & 155.37 & 155.37 & 63.38\\
6 & 64.5 & 64.5 & 211.79
\end{tabular}
\end{table}

\subsection{Functional data} To check the robustness of our proposal, we now consider a set of $n$ = 100 functions that are generated from the following model, which was analyzed previously by  \cite{ENV:ENV878}, \cite{FRAIMAN2013326} and \cite{Arribas} for functional outlier detection procedures. $n - \left\lceil cr \cdot n\right\rceil$ are generated from $X(t)$ =  $30t(1-t)^{3/2} + \epsilon(t)$, whereas the remaining $\left\lceil cr \cdot n\right\rceil$ functions are generated from this contamination model: $30t^{3/2} (1-t) +  \epsilon(t)$, where $t \in \left[ 0, 1 \right]$ and $\epsilon(t)$ is a Gaussian process with
zero mean and covariance function  $\gamma (s,t)$ = $0.3 \exp \{ - \left| s - t \right| /0.3\}$. The functions are measured at 50 equispaced points between 0 and 1. A total of 100 simulations have been run with two contamination rates $cr$ = 0.1 and 0.15. Original and robust ADA have been applied with $k$ = 2 archetypoids. With $cr$ = 0.1, 10\% of the times one outlier belongs to the solution for original ADA, while no outlier is included as an archetypoid for robust ADA. With $cr$ = 0.15, 78\% of the times one outlier belongs to the solution for original ADA, while this percentage was only 32\% for robust ADA. Therefore, our proposal provides robust solutions.

{Let us analyze the influence that $c$ has on the results. Table \ref{cout} shows the percentage of times one outlier belongs to the robust ADA solution for different $c$ values. It seems that, for these data, a multiplicative factor of 1 gives more robust results than if we used a multiplicative factor of 6. Nevertheless, even with a multiplicative factor of 6 and for all the  percentiles considered, the results are more robust than those obtained using the original ADA.}

\begin{table}  
\caption{{Percentage of times one outlier belongs to the robust ADA solution for different $c$ and $cr$.} \label{cout}}
\centering
\begin{tabular}{lrrrrrr}
 & \multicolumn{3}{c}{$cr$ = 0.1} & \multicolumn{3}{c}{$cr$ = 0.15} \\
Multiplicative & \multicolumn{3}{c}{Percentile} & \multicolumn{3}{c}{Percentile}\\
Factor & 25th & 50th & 75th & 25th & 50th & 75th\\ \hline 
1 & 0 & 0 & 0 & 0 & 1 & 2\\
6 & 0 & 0 & 2 & 8 & 32 & 52
\end{tabular}
\end{table}

\begin{figure}
\begin{center}
\begin{tabular}{cc}
	\includegraphics[width=.5\linewidth]{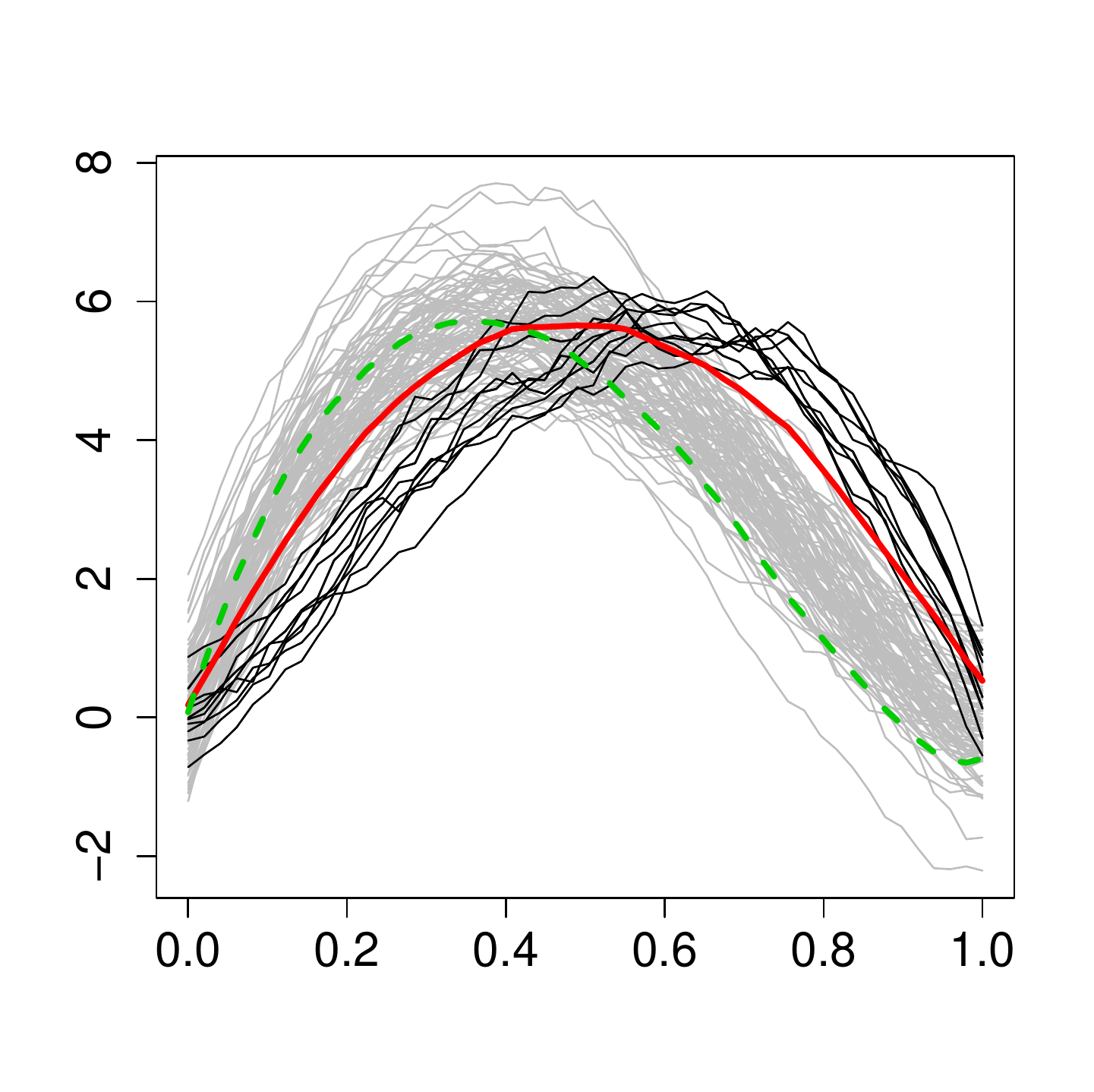} & \includegraphics[width=.5\linewidth]{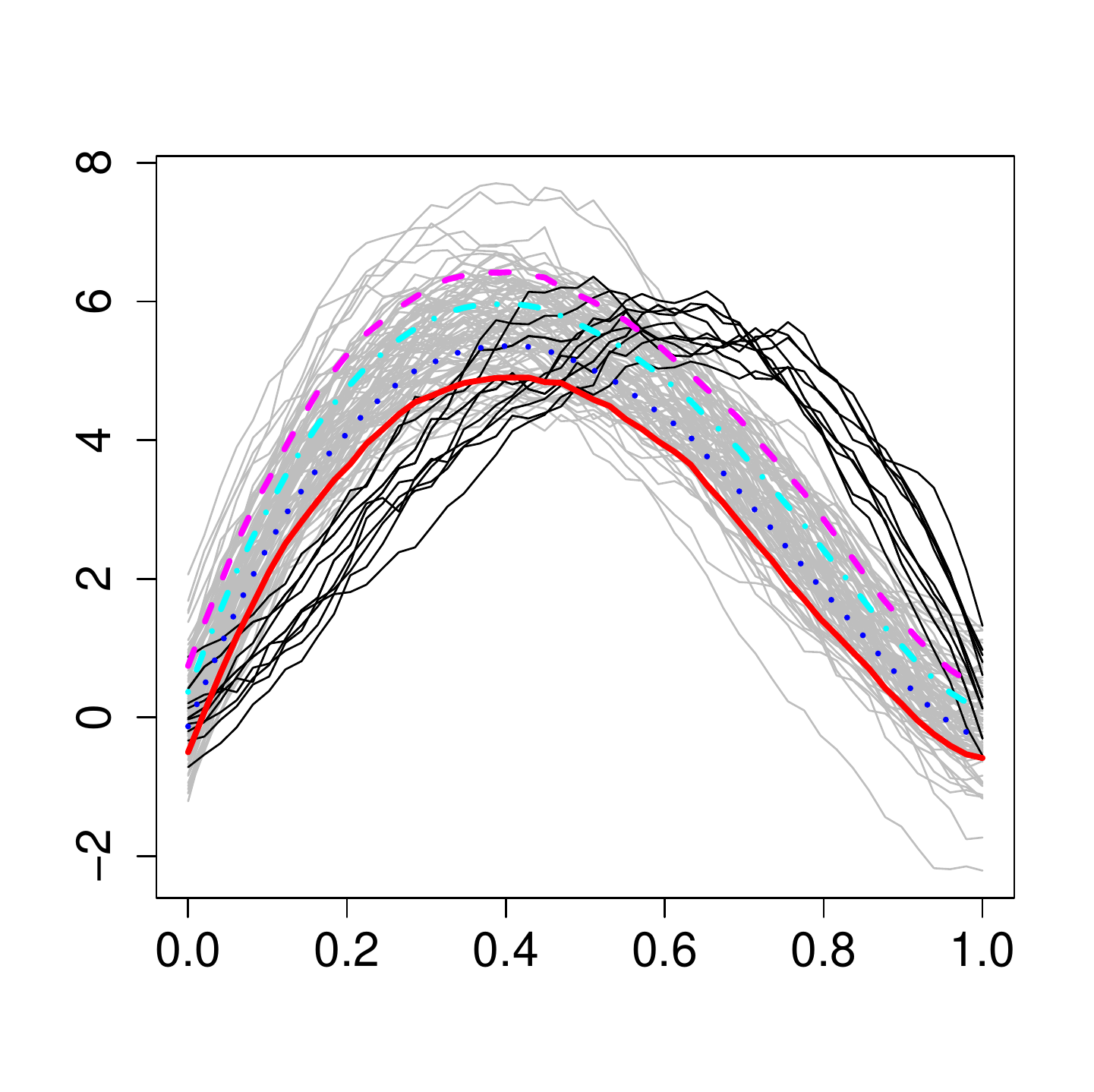} 
 \\
	(a) & (b) \\
	\includegraphics[width=.5\linewidth]{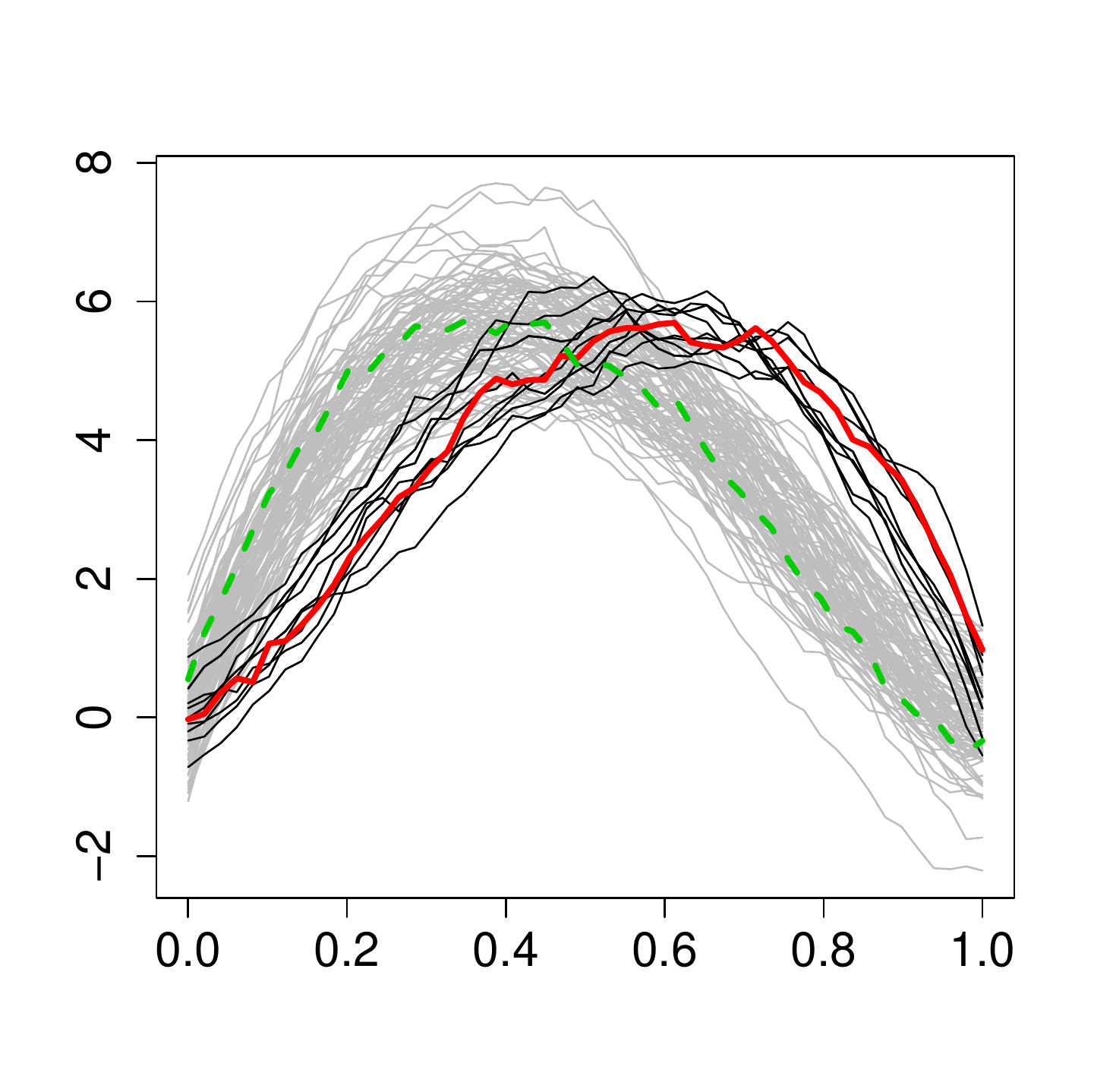} & \includegraphics[width=.5\linewidth]{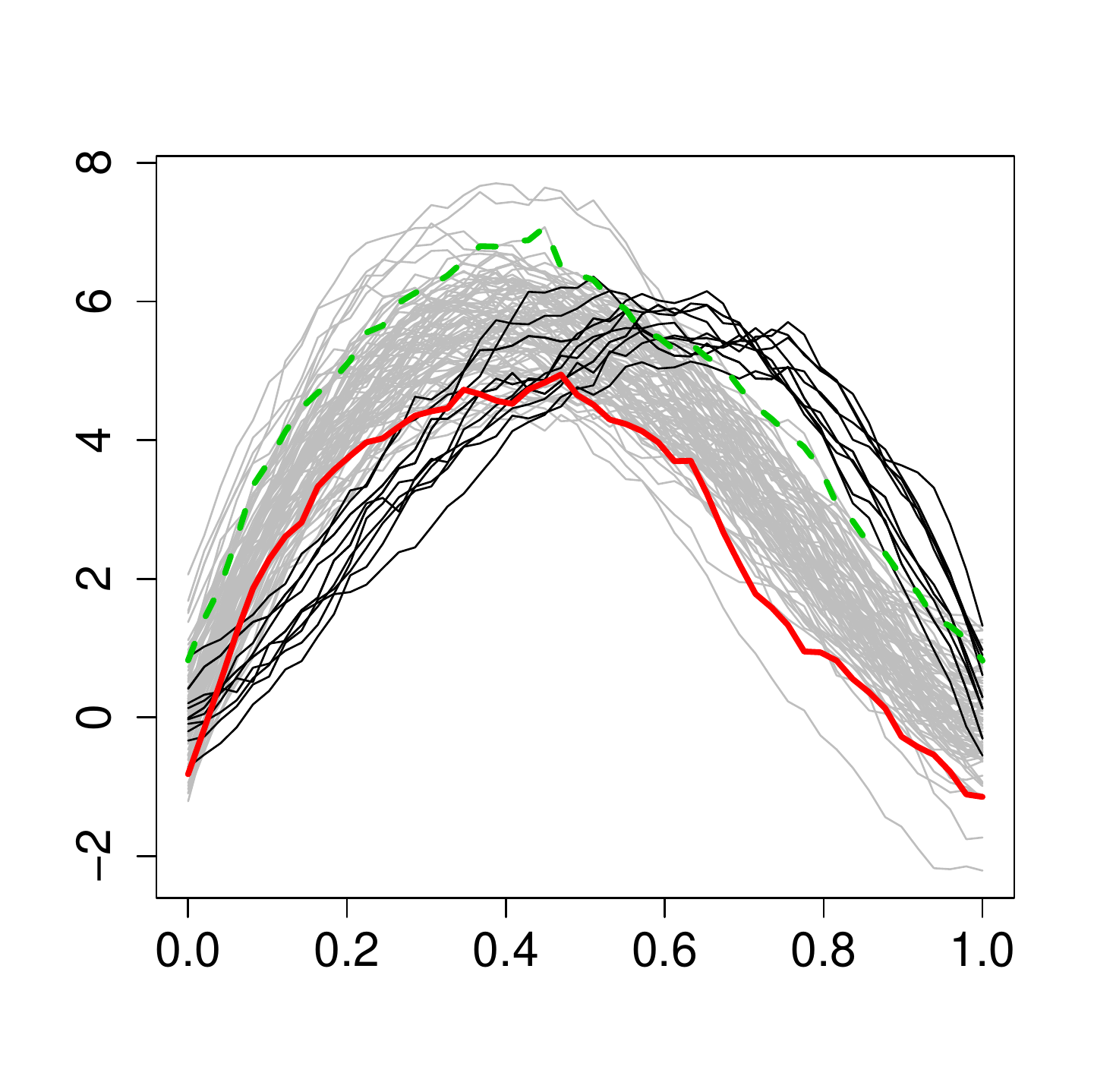} 
	 \\
	(c) & (d) 
	\end{tabular}
	\end{center}
	\caption{{Simulated model with 10 outliers. Data are shown in gray  and outliers in black. (a) Effect of adding and subtracting a multiple of PC 1 to the mean curve. (b) Effect of adding and subtracting two different multiples of robust PC 1 to the center curve (the multiple is 2 for blue functions and 5 for red and purple functions). (c) Archetypoids with the original ADA algorithm. (d) Archetypoids with the proposed robust ADA algorithm.}
	\label{adapcafig}}
\end{figure}

{Let us visually compare the solutions obtained for one of the simulation with $cr$ = 0.1, i.e. with 10 outliers, when an outlier is selected as an archetypoid by the original ADA algorithm. Remember that no outlier is selected as an archetypoid with our robust proposal. In Figure \ref{adapcafig}, we compare those solutions with the solutions obtained by PCA and robust PCA, as developed by \cite{doi:10.1198/004017004000000563} and \cite{Engelen_Hubert_Vanden_Branden_2016} and implemented in the function $robpca$ from the R package {\bf rospca} \citep{rospca}.  PC 1 is nearly zero in the first half of the interval, but 0.2 in the second half of the interval. This result is highly influenced by the outliers. The two archetypoids with the original ADA return a similar base as with PCA. However, the robust versions of PCA and ADA return results that are similar to each other, but different from their respective non-robust versions. In both cases, the robust version returns solutions that are like a band \citep{doi:10.1198/jasa.2009.0108} of the non-contaminated data. The robust PC 1 is nearly constant (-0.15) along  the entire interval. Depending on the multiple considered  the band covering the data is more or less wide.}

{The previous comparison is qualitative. To make a quantitative comparison we can take into account that outliers can be detected by their large deviation from the robust fit, as obtained with the function $robpca$. For the ADA methodology, we propose to compute the robust archetypoids with $c$ = $P_{50}$ because it is conservative and ensures a more robust solution without outliers as archetypoids. Then, we compute $||r_i||$ and a box plot is applied to this distribution to detect the outliers. We call this methodology RADAB. Furthermore, we also consider the method by \cite{HYNDMAN20074942} (ISFE), who use  integrated square forecast errors and robust principal component analysis to detect outliers, with the function $foutliers$ and the option $HUoutliers$ from the R package {\bf rainbow} \citep{rainbow}. Table \ref{medidasoutliers} shows the results (True Positive Rate, TPR, False Positive Rate, FPR, and Matthews correlation coefficient, MCC) for different $cr$ values. $robpca$ has the highest TPR, but at the expense of having the highest FPR and the lowest MCC. With no outlier ($cr$ =0), ISFE has the lowest FPR, but FPR is smaller for RADAB with $cr$ = 0.1 and $cr$ = 0.15. On the other hand, RADAB reports excellent results with TPR, with nearly 100\% for TPR, which is not the case with ISFE. In fact, the maximum for MCC is achieved with RADAB in all the cases. }

\begin{table}  
\caption{{Mean and standard deviation, in brackets, of TPR (percentage), FPR (percentage) and MCC for different $cr$.} \label{medidasoutliers}}
\tiny
\centering
\begin{tabular}{c@{\hspace{0.3\tabcolsep}}c@{\hspace{0.3\tabcolsep}}c@{\hspace{0.3\tabcolsep}}c@{\hspace{0.3\tabcolsep}}c@{\hspace{0.3\tabcolsep}}c@{\hspace{0.3\tabcolsep}}c@{\hspace{0.3\tabcolsep}}c@{\hspace{0.3\tabcolsep}}c@{\hspace{0.3\tabcolsep}}c@{\hspace{0.3\tabcolsep}}c@{\hspace{0.3\tabcolsep}}}
 Method &  $cr=0$ & \multicolumn{3}{c}{$cr$ = 0.05} & \multicolumn{3}{c}{$cr$ = 0.1} & \multicolumn{3}{c}{$cr$ = 0.15} \\
& FPR & TPR & FPR & MCC & TPR & FPR &  MCC & TPR & FPR & MCC\\ \hline
$robpca$ & 10.75 (2.94) & 100 (0) &     8.89  (2.69) & 0.59 (0.06) &  100 (0) & 7.69 (2.36) & 0.74 (0.06) & 100 (0) & 5.94 (2.08) & 0.84 (0.05)\\  
RADAB & 4.82 (2.14) & 99.40 (4.45)  &  3.00 (1.98) & 0.80 (0.10) & 99.0 (7.18) & 1.58 (1.66) & 0.93 (0.09) & 97.93 (10.58) & 0.52 (0.89) & 0.98 (0.03)\\
ISFE & 2.99 (2.0) & 86.00 (29.75) & 2.94 (1.84) & 0.70 (0.26)& 84.90 (30.99) & 2.71 (1.77) & 0.78 (0.26) & 77.13 (33.10) & 2.41 (1.85) & 0.75 (0.29)
\end{tabular}
\end{table}


\section{Application}
\label{resultados} When dealing with time series, the theoretical complexity of many of the statistical methods available for analysis 
leads to periodic summaries of the data series being commonly used in practice. Furthermore, many of the techniques, such as the classic Box-Jenkins theory (\cite{box1976time}), involve verifying  a set of quite restrictive hypotheses, such as stationarity, equally-spaced observations or belonging to a specific kind of well-known processes. What makes  our proposal attractive is not only the lack of these restrictive hypotheses (it could also be applied to sparsely measured time series), but also the data speak for themselves and the results can be interpreted easily by non-specialists. It also allows them to be visualized, which is an important task (\cite{JSSv025c01}).
To illustrate these claims, robust bivariate FADA is applied to the company stock prices as detailed below.

\subsection{Data}
We consider a data set from \cite{quantcuote2017}, which is composed of a collection of daily resolution data with the typical open, high, low, close, volume (OHLCV) structure. This collection runs from 01/01/1998 to 07/31/2013 for 500  currently active symbols in the S\&P 500. In addition, the aggregate S\&P 500 index OHLCV daily tick data (\cite{yahoofinance2017}) have been  added to the data set.  The length of this time series has been selected so that its range coincides with that of the QuantQuote disaggregated series. We  have also collected information from 
\cite{sectorSPDR2017}, which classifies stocks on the S\&P 500 index within ten major sectors, namely:
Consumer Discretionary (CD),
Consumer Staples (CS),
Energy (E),
Financials (F),
Health Care (HC),
Industrials (I),
Materials (M),
Real Estate (RE),
Technology (T), and 
Utilities (U).

We are interested in extracting financially relevant information. With regard to investments, and more specifically to portfolio theory, it is widely accepted that the two key variables are risk and profitability (\cite{markowitz1952portfolio}). On the one hand, we have chosen as a measure of profitability in a time $t$ the aggregate returns over a period of length N, {$r_N$}. So, for each price series, $s_i,\, i=1,...,501$ ${r_N}^{s_i}(t)=\frac{X^{s_i}_{t}-X^{s_i}_{t-N}}{X^{s_i}_{t-N}}$
where $X_t^{s_i}$ is the value of the stock $i$ at the time $t$. In our case, we have chosen $N$ = 250, approximately the number of days that stock markets remain open in a year. On the other hand, we have chosen as a measure of volatility the beta or $\beta$ coefficients, which are widely used in portfolio theory.  This is defined for a particular stock $s_i$ in a time $t$ as	
$\beta_N^{s_i}(t)= \frac{Cov({r_N}^{s_i}(t),{r_N}^{index}(t))}{Var({r_N}^{index}(t))}$
where ${r_N}^{s_i}(t)$ stands for the aggregated returns of $s_i$ in time $t$ over the last $N$ days and ${r_N}^{index}(t)$  are the returns of the aggregated S\&P 500 index in the same period. To be consistent, we perform the calculations with a time frame of 250 days. 

Although FDA can handle missing data well, as explained previously, our sample presents a different problem: some stocks do not exist throughout the entire time series. This is not because we have missing data, but because the companies were founded or were first listed on the stock exchange after 01/01/1998, i.e. the domains of the functions are different. Here we propose an alternative  to deal with this problem, trying to maximize the size of our sample and minimize the stocks that must be discarded by taking into account observations since 2000-01-01 and dropping the stocks with more than 20\%  missing values. In this way, only four companies from 500 are left out.

The following step is to transform discrete data to functional data. The {cubic B-spline} basis {with equally spaced knots} has been considered because the series are {not periodic}. As regards the number of bases $m$, as suggested by \cite{Ramsay05}, we computed an unbiased estimate of the residual variance using 4 to 22 bases
and selected $m$ = {13}, which is the number of bases that makes decreasing the residual variance substantially. In summary, for each stock $s_i$, both variables, return and beta coefficient in a 250 day time frame, could be expressed as the functions: $r_{250}^{s_i}(t)=\sum_{h=1}^{{13}}\mathbf{a}_h^{s_i}\phi_h(t)$ and $\beta_{250}^{s_i}(t)=\sum_{h=1}^{{13}}\mathbf{b}_h^{s_i}\phi_h(t)$ where $\mathbf{a}^{s_i}$ stands for the vector of coefficients on the basis functions for $r_{250}(t)$ corresponding to the particular stock $s_i$ and, in the same way, $\mathbf{b}^{s_i}$ stands for the vector of coefficients for $\beta_{250}(t)$ corresponding to the particular stock $s_i$, and $\phi_h$ are the {B-splines} basis functions. Therefore, after smoothing, the original data set of dimensions $496 \times 3422 \times 2$ is reduced to $496  \times {13} \times 2$.
As both functions are measured
in non-compatible units, each functional variable is standardized before  analysis by standardizing the coefficients in the basis  as explained by \cite{Epifanio2016}. This data set was analyzed in a non-robust way by \cite{MolEpi18}{, using the Fourier basis}.

\subsection{Robust bivariate FADA}
Table \ref{tb:ArchetypoidTable} shows the companies obtained as archetypoids for different $k$ values. When $k$ increases, the number of sectors represented in the set of archetypoids increases too. Since archetypoids are not nested, archetypoids may not coincide at all when the $k$ value varies. However, we find a nested structure in the order in which sectors appear, and in fact, some companies remain when $k$ increases, such as {AKAM or XLNX}.

\begin{table} 
\centering
\caption{Archetypoids for different $k$ values. We use the same symbols for company names as \cite{sectorSPDR2017}.  The abbreviation of the economic sector to which each company belongs appears in parentheses. \label{tb:ArchetypoidTable}} 
\begin{tiny}
{
\begin{tabular}{lllllllllllll}
$k$ & A1 & A2 & A3 & A4 & A5 & A6 & A7 & A8 & A9 & A10 \\
3 &ALTR (T) & LNC (F)       & WEC (U) &                   &                &                   &              &                       &                         &                 \\
			4  & ALTR (T) & LNC (F)       & NUE (M) & GIS (CS)   &                &                   &              &                       &                         &                 \\
			5  & ALTR (T) & HIG (F)       & NUE (M) & GIS (CS)   & KIM (RE) &                   &              &                       &                         &                 \\
			6  & BRCM (T) & IPG (CD) & NUE (M) & GIS (CS)   & KIM (RE) & DNR (U)   &              &                       &                         &                 \\
			7  & AKAM (T) & C (F)         & WMT (CS) & HRL (CS)   & BXP (RE) & TER (T)  & SWN (E) &                       &                         &                 \\
			8  & AKAM (T) & RF (F)        & NUE (M) & GIS (CS)   & EQR (RE) & XLNX (T) & NBR (E) & CVC (CD) &                         &                 \\
			9  & AKAM (T) & MS (F)        & X (M)   & SO (U)    & AIV (RE) & XLNX (T) & SWN (E) & IPG (CD) & BRKB (F)        &                 \\
			10 & AKAM (T) & MS (F)        & ATI (M) & FLIR (T) & KIM (RE) & XLNX (T) & EOG (E) & GCI (CD) & CMCSA (CD) & GIS (CS)   \\
\end{tabular}}
\end{tiny}
\end{table}

In the interests of brevity and as an illustrative example we analyze the
results of $k$ = {4}, which is the value selected by the elbow criterion. We include a brief description of the data-driven selected companies based on the information in \cite{sectorSPDR2017}: {Altair Engineering Inc. (ALTR), together with its subsidiaries, provides enterprise-class engineering software worldwide; Lincoln National Corp.(LNC) is a holding company. Through subsidiary companies, the company operates multiple insurance and investment management businesses; Nucor Corporation (NUE) manufactures and sells steel and steel products in the United States and internationally; General Mills, Inc. (GIS) is a leading global food company. Its brands include Cheerios, Annie's, Yoplait, Nature Valley, Fiber One, Haagen-Dazs, Betty Crocker, Pillsbury, Old El Paso, Wanchai Ferry, Yoki and others.}

\begin{figure}
\begin{center}
	\includegraphics[width=.7\linewidth]{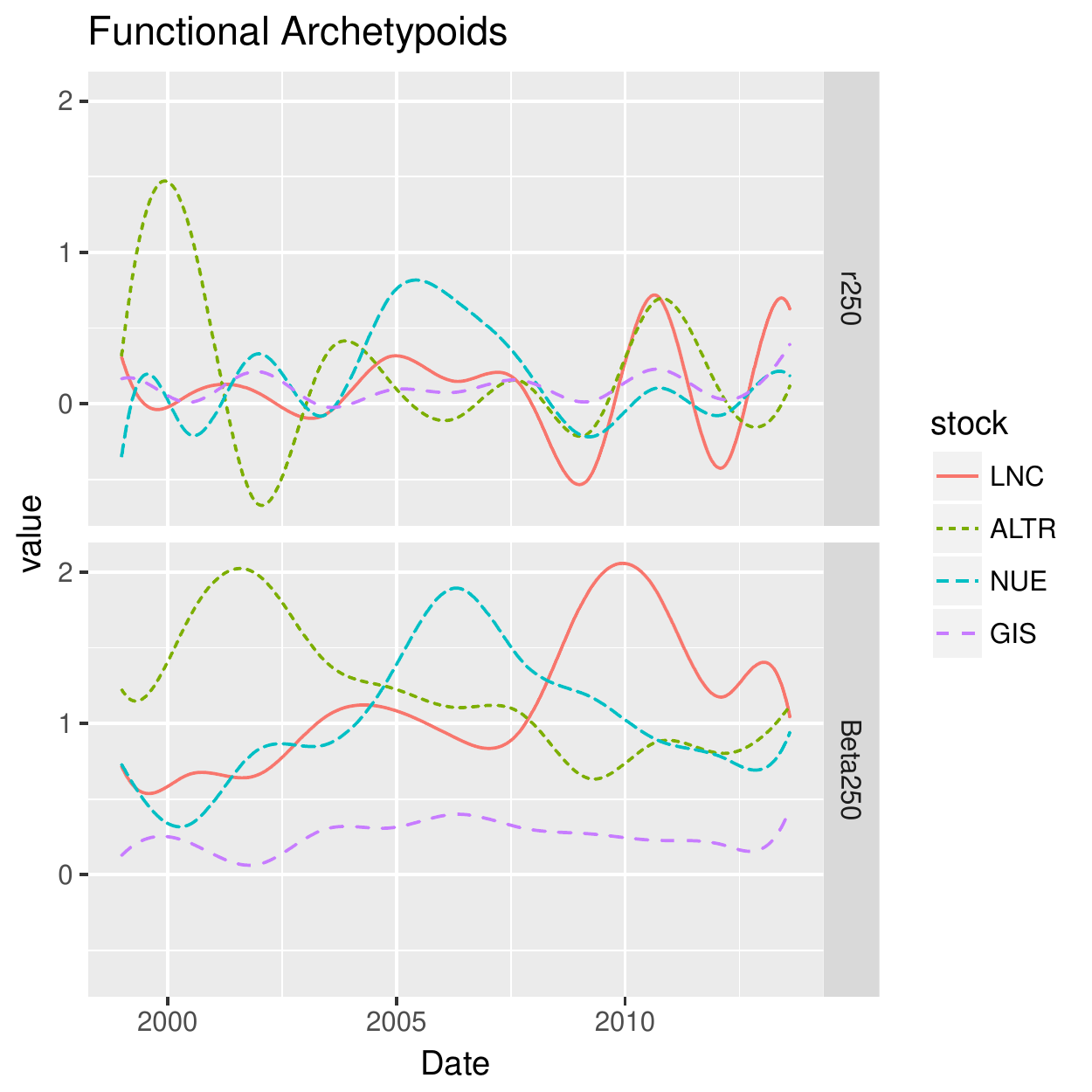}
	\end{center}
	\caption{$r_{250}$ and $\beta_{250}$ functions of the {4} archetypoids.
	\label{fig:archoverlaped}}
\end{figure}

Figure \ref{fig:archoverlaped} shows the functions of each variable for the {4}  archetypoids.
{It can be seen that GIS is a company that, in comparison with the rest of the archetypoids, presents low and constant values for both variables. Looking at ALTR,  it presents the highest returns and volatilities at the beginning of the time series probably as a result of the .com bubble. LNC presents the typical profile of a financial company, with moderate profitability and volatility during the first three quarters of the time series. Once the crisis broke out in 2007, volatility shot up to unprecedented levels while profitability plummeted. Finally,  NUE  is characterized by having bell-shaped functions, that is, with relatively low values at the extremes of the temporal domain and higher values at the center.}

What makes this technique very interesting is that the companies are expressed as mixtures of those extreme profiles through the $\alpha_{ij}$ values. For a visual overview of the S\&P 500 stocks, we propose the following procedure that provides a data-driven taxonomy of the S\&P 500 stocks. The idea is to group together companies that share similar $\alpha_{ij}$  profiles. A simple method to do this is to establish a threshold $U$, so that if the weight of an archetypoid for a given individual is greater than $U$, we will say that this subject is in the cluster generated by this archetypoid. If we repeat this process for each archetypoid, we will generate {4} ``pure" clusters of subjects, i.e., clusters of subjects that are represented mostly by a single archetypoid. To take this a little further, this process is repeated with combinations of two archetypoids. Thus,	 we will group in the same cluster the individuals whose sum of {$\alpha$} weights  for two concrete archetypoids is greater than $U$, thus generating ${\binom{4}{2}=6}$ additional clusters.  In this case, it might happen that for a given subject, there is more than one combination of archetypoids whose sum exceeds $U$. So as not to complicate the graphic representation, we will classify these subjects generically as mixtures, even though these mixtures will be composed of different sets of archetypoids.{Note that we do not carry out a clustering method, but we form clusters according to the alpha values.}

\begin{sidewaysfigure}
	\centering
	\includegraphics[width=1\linewidth]{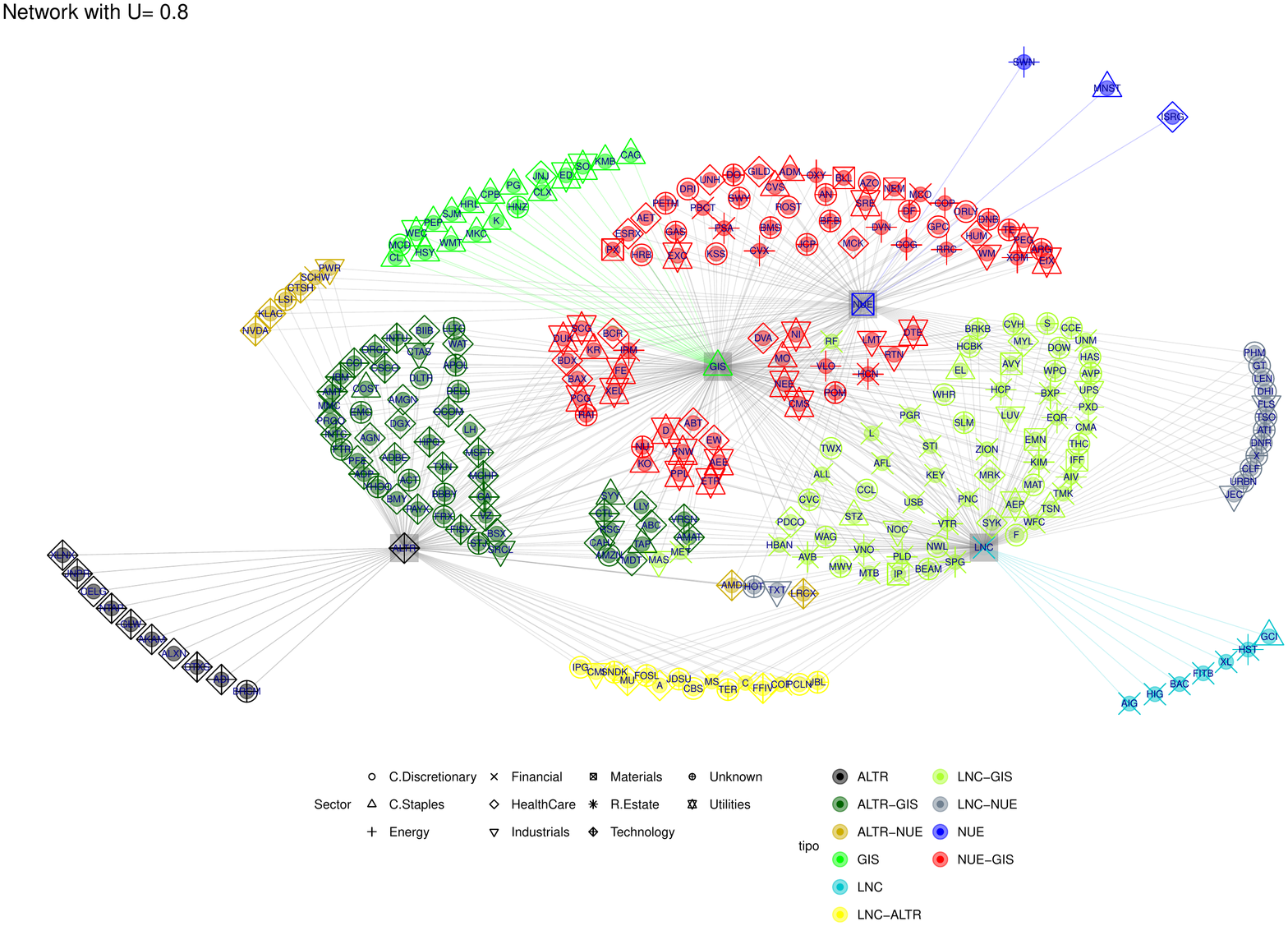}
	\caption{Cluster structure with $U$=0.8. Sector indicates the economic sector and tipo the type of cluster. \label{fig:red80r}}
\end{sidewaysfigure}

Figure \ref{fig:red80r}  condenses all the information extracted by means of a network that is built through an adjacency matrix that gathers the cluster information.   The smaller  $U$ is, the larger the number of companies that belong to any cluster, and vice versa. We have chosen a value of $U$ = 0.8, which is high but not so much, i.e. $U$ = 0.8 is a breakeven between having a considerable number of companies, but at the same time not too many to be able to read the graphic representation concisely. Archetypoids are highlighted with a gray square, the lines and color codes allow us to differentiate the structure of the clusters, and  different sectors are represented through different geometrical shapes.

{Starting with the pure clusters we see that NUE generates a small heterogeneous group with 3 other companies. 
GIS is the archetypoid that generates the largest cluster on its own. Most of the companies in this cluster belong to the C. Staples sector, although we also find some from the Utilities sector. 
LNC generates a small cluster with seven other companies. Five of them, Bank of America Corp (BAC) and Hartgord Financial Services Group INC. (HIG), American International Group Inc (AIG), Fifth Third Bancorp (FITB ) and XL Group Ltd (XL) belong to the same sector and have similar profiles. The other two are  Gannett Co. Inc. (GCI) related to the publishing industry and Host Hotels and Resorts Inc (HST), a real estate investment trust. 
For its part, ALTR generates a larger cluster in which 10 other companies are included. Most of these companies belong to the Technology sector. }

{
Regarding mixed clusters, we will point out some general characteristics, since detailing all the relationships shown in Figure \ref{fig:red80r} would be too extensive. 
 NUE and GIS form the largest cluster for this level of $U$. Focusing on this, the NUE-GIS cluster has just a few companies from the C. Staples sector, since these companies are mainly classified in the cluster generated by GIS alone. Many of the companies classified in the NUE-GIS cluster belong to the Utilities,  Energy and Health sectors. 
 It could be said that the common feature of the companies in this cluster is that they are not very sensitive to the economic cycle.}

{
The following most important clusters according to their size are LNC-GIS and ALTR-GIS.  The LNC-GIS cluster is certainly heterogeneous in terms of the sectors that comprise it.
However, it can be seen that the majority of Financial and Real Estate companies are part of this group.}

{
The third largest cluster is formed by ALTR-GIS and is quite homogeneous. The majority of the companies in this group belong to the technology sector. If we look at the other two clusters in which ALTR intervenes we see that this feature is repeated. Both ALTR-NUE and ALTR-LNC have a large presence of technology companies and, what is more, we do not find technological companies outside these groups.}

{
The last cluster is formed by LNC-NUE. It is formed by 14 companies and the sector that appears the most times is the C. Discretionary sector.
}

\begin{figure}
	\centering
	\includegraphics[width=0.8\linewidth]{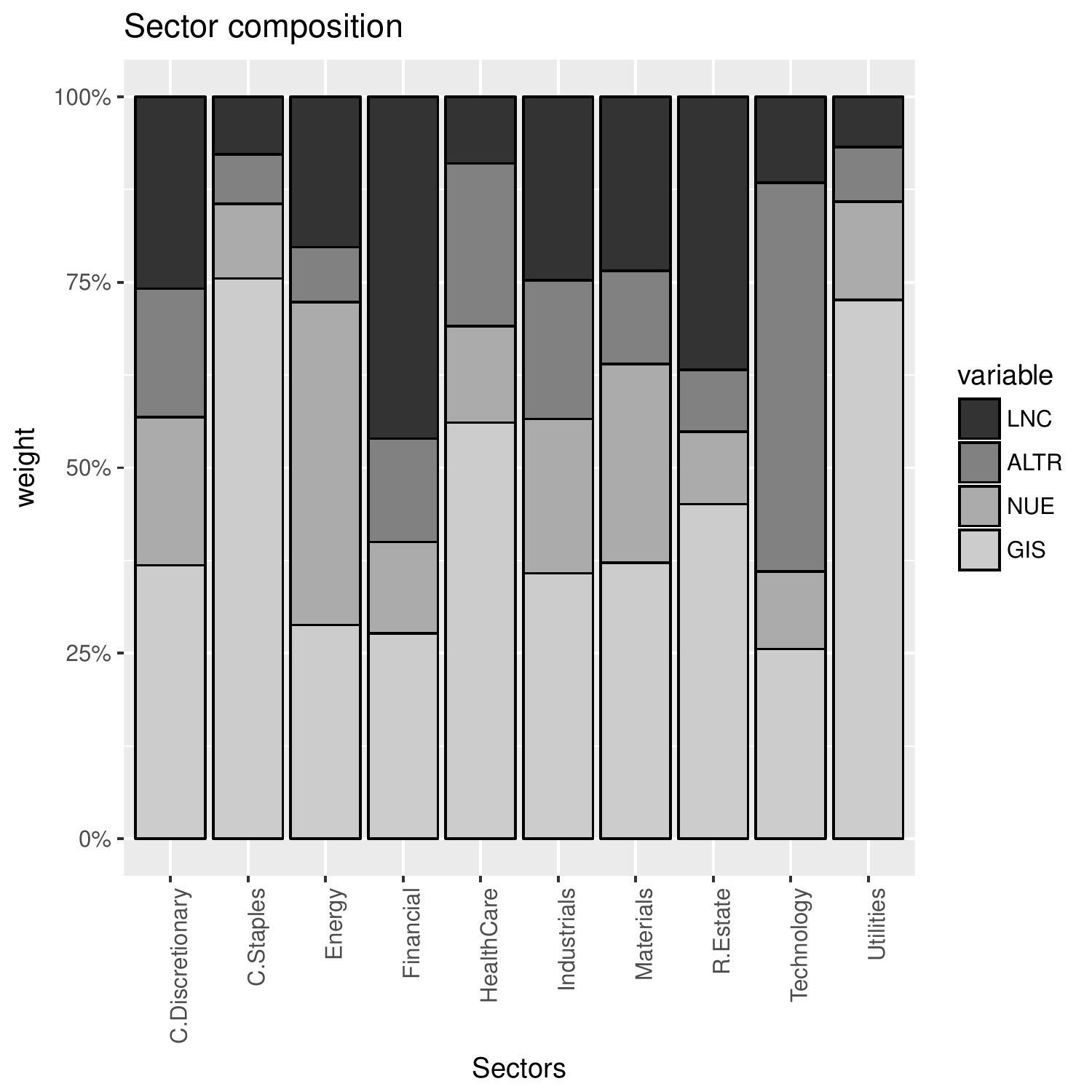}
	\caption{Relative weight of archetypoids in each sector.\label{fig:archbysector}}
\end{figure}

Now we analyze the composition of the sectors managed by market analysts to evaluate the performance of the results qualitatively. Figure \ref{fig:archbysector} shows the normalized relative weight of archetypoids in each of the ten sectors. {It can be seen that each archetypoid represents the component with the greatest weight of the sector to which it belongs. Thus, LNC accounts for more than 45\% of the financial sector, GIS represents almost 80\% of the weight of the C. Staples sector and in the technology sector, the weight of ALTR exceeds the weight of the other three archetypoids. 
 }

{But it is not only interesting to analyze the weights of the most relevant components. The composition of the mixtures also gives clues about the similarities and dependencies between the different sectors. For example, if we compare Consumer Discretionary sector with Consumer Staples sector, we see that  the weights keep a proportion that we would expect. For example, the companies that manufacture durable goods, which are included in the Consumer Discretionary sector, have a direct relationship with those that provide the investment to finance these purchases  represented by the LNC archetypoid, which is why that archetypoid has higher weight in this sector. 
On the other hand, companies that provide basic or non-durable goods, which belong to the Consumer Staples sector, have a minor relationship with the financial sector. It may be obvious, but it is worth emphasizing that, by definition, non-durable goods are those that are purchased without resorting to financing. Instead, this sector has a great similarity with the Utilities sector (distributors of electricity, water, gas, etc.). This makes economic sense, since basic goods and services distributed by companies in the Utilities sector have similar demand curves. In other words, in the expansive cycles of the economy, consumers decide to increase their investments in goods that require financing such as a car, a washing machine or a computer. However, households' spending on electricity, water, gas or telecommunications will remain relatively constant, as will spending on basic products such as bread, milk, oil, soap and toothpaste.
}

{
Regarding the composition of the energy sector (extractors of oil, gas etc), the small weight of the technological archetypoid is noteworthy. This may point to the weak relationship between the technologies used in each sector. On the one hand, the Energy sector carries out activities where  heavy machinery and  mechanical operations in general (drilling, mining, extraction etc.) are fundamental. This is completely opposite to the dynamics that prevail in the technological universe, where the main elements are computer applications, digital technology and patents.}

{The Industrials and Materials sectors have similar profiles, which shows the strong interrelation between them. Regarding the Real Estate sector, we see how the LNC archetypoid, belonging to the financial companies sector, has the greatest weight outside its own sector. The relationship between these two sectors is also evident. Finally, we can say that the Utilities and C. Staples sectors are in some way the purest, in the sense that the weight of the dominant archetypoid of this sector is greater than the weight of the other three archetypoids all together. 
}


\section{Conclusion}
\label{conclusiones}
This paper introduces robust archetypal analysis (AA and ADA) for multivariate data and functional data, both for univariate and multivariate functions. A simulation study has demonstrated the good performance of our proposal. 

Furthermore, the application to the time series of stock quotes in the S\&P 500 from 2000 to 2013 has illustrated the potential of these unsupervised techniques as an alternative to the commonly used clustering of time series, which are usually described by complex models. 
Understanding the results of these or other statistical learning models is not always an easy task. Additionally, if these results have to be explained to a public with little or no  mathematical knowledge, things can be even worse. In this regard, ADA is particularly suitable  since its results can be interpreted in a very simple way by any non-expert person. For instance, anyone with minimal knowledge of investments  understands what we mean if we say that a certain stock behaves like a mixture of {Nucor Corporation and Lincoln National} stocks. Another advantage of the applied model is that the functional version of archetypoid analysis (FADA) allows us to condense vectors of observations of any length into a few coefficients, which provides an improvement in computational efficiency and makes this method highly recommended when working with long time series.

With regard to the financial conclusions, in the first place, it has been seen that when we increase the number of chosen archetypoids, the Technology sector appears repeatedly while other sectors do not appear. Therefore, there are companies within this sector that exhibit very different behaviors, such as {XLNX}, FLIR and AKAM.

Furthermore, we have proposed a visual representation of the companies through the definition of clusters based on the mixtures obtained. 
Finally, we have analyzed the sectors according to the normalized relative weight of archetypoids that compose each sector and we have seen that some sectors present certain similarities. It is worth mentioning that  sectors like Consumer Discretionary, Materials or Industrials offer better opportunities for diversifying risks, since their composition is more heterogeneous. On the other hand, sectors such as Utilities or Consumer Staples present more homogeneous structures, where the weight of the dominant archetypoids of the sector can exceed {70\%}.

As regards future work, from the mathematical point of view, {an open question, both in the real and functional case, is the selection of the tuning parameter of the loss functions. The research line of using robust archetypal analysis for outlier detection is another open path. Another} open problem is the extension of archetypal analysis to mixed data (functional and vector parts). An appropriate definition of the inner product  is needed {since functional and vector parts will not be measured in directly compatible units}. Nevertheless, applications, even beyond econometrics, are the main direction of  future work. Even so, the application of these models to the world of finance  is still a relatively unexplored field. The application of models with functional data allows us to take into account variables collected with different frequencies, such as daily quotes, quarterly balances or annual results, which makes these models especially suitable for financial time series.

Taking this into account, a future development may be to extend the implementation of the bivariate model to a $P$-variable model that makes it possible to work with a large amount of data from each company. 	From a financial point of view, it may be possible to develop investment strategies using the results shown here to improve performance and reduce the risk of investment decisions. {In fact, an interesting open problem to study is whether the estimated archetypoids may be useful for constructing small portfolios of stocks or a variety
of small (i.e., $k$ stock) portfolios.}

\section{Acknowledgements}
{The authors are grateful to the Editor and  reviewers for their very
constructive suggestions, which have led to improvements in the manuscript.}
This work is supported by DPI2017-87333-R from the Spanish Ministry of Economy and Competitiveness (AEI/FEDER, EU) and UJI-B2017-13 from Universitat Jaume I. 



\end{document}